\documentclass[conference]{IEEEtran}
\IEEEoverridecommandlockouts

\usepackage{cite}
\usepackage{amsmath,amssymb,amsfonts}
\usepackage{graphicx}
\usepackage{tipa}
\usepackage{subcaption} 
\usepackage{textcomp}
\usepackage{xcolor}
\usepackage{url}

\title{Understanding When and Where Self-Supervised Speech Models Learn Phonemes and Prosodic Cues through Layer-Wise and Time-Resolved Probing\\

}
\author{
\IEEEauthorblockN{
Masaru Tanibata,
Shun Takahashi,
Hiroki Ouchi,
and Sakriani Sakti
}
\IEEEauthorblockA{
Division of Information Science, Nara Institute of Science and Technology, Japan\\
\texttt{\{tanibata.masaru.tr2,takahashi.shun.tq9\}@naist.ac.jp} \\
\texttt{\{hiroki.ouchi,ssakti\}@is.naist.jp}
}
}

\begin{document}

\makeatletter
\renewcommand{\@IEEEpubidpullup}{3\baselineskip}
\makeatother
\IEEEpubid{%
  \begin{minipage}[b]{\textwidth}
  \scriptsize\raggedright
  \copyright\ 2026 IEEE. Personal use of this material is permitted.
  Permission from IEEE must be obtained for all other uses, in any current or
  future media, including reprinting/republishing this material for advertising
  or promotional purposes, creating new collective works, for resale or
  redistribution to servers or lists, or reuse of any copyrighted component
  of this work in other works.
  \end{minipage}%
}

\maketitle

\begin{abstract}
Self-supervised speech models (S3Ms) encode a rich hierarchy of acoustic,
prosodic, and phonetic information, yet \emph{when} and \emph{where} it emerges
during pretraining remains unclear. We train frozen-encoder probes on every
Transformer layer across pretraining checkpoints of two matched models:
HuBERT-Base, whose stage~2 is a separately initialized model trained on new
pseudo-label targets, and single-objective wav2vec 2.0-Base. Our five tasks span
an acoustic-to-abstract gradient: absolute pitch (F0), speaker-relative pitch,
lexical stress, phoneme boundaries, and 38-way phoneme identity. Acoustic--prosodic cues
and phone boundaries are decodable from the earliest checkpoints and barely
shift across layers, whereas phoneme identity is constructed during pretraining,
re-localizing to upper-middle layers under HuBERT's stage-2 targets at a matched
update count. Acquisition order therefore tracks abstraction from the
acoustic surface, not the segmental/suprasegmental distinction, extending to
prosody an account established for phonetic, lexical and syntactic structure;
the no-switch wav2vec 2.0 shows no comparable re-localization.
\end{abstract}

\section{Introduction}

Self-supervised speech models (S3Ms) learn representations directly from raw
audio, without human-provided labels, by solving pretext tasks such as
contrastive prediction over masked latent
features\cite{baevskiWav2vec20Framework2020} or masked prediction of
self-generated pseudo-labels\cite{hsuHuBERTSelfSupervisedSpeech2021}. Trained
on large amounts of unlabeled speech, they have become foundational components
of modern speech
technology\cite{aroraLandscapeSpokenLanguage2025,mohamedSelfSupervisedSpeechRepresentation2022},
yet the internal representations they acquire during pretraining remain only
partially understood.

Although optimized for low-level acoustic objectives, S3Ms encode a rich linguistic hierarchy: phonetic and phonemic distinctions\cite{wellsPhoneticAnalysisSelfsupervised2022,seysselProbingPhonemeLanguage2022a,
martinProbingSelfsupervisedSpeech2023a,
choiSelfSupervisedSpeechRepresentations2024}, word
identity\cite{pasadWhatSelfSupervisedSpeech2024}, and phonological structure\cite{venkateswaranProbingPhonologySelfSupervised2025}. This information is distributed systematically across depth, with acoustic and speaker cues in the low layers,
phonetic and higher-level attributes in the middle, and a re-specialization
toward the pretraining objective at the
top\cite{pasadLayerwiseAnalysisSelfsupervised2022,
pasadComparativeLayerwiseAnalysis2023b,chungSimilarityAnalysisSelfSupervised2020,
liuSelfsupervisedPredictiveCoding2023,mohamedOrthogonalityIsotropySpeaker2024,
gubianAnalyzingRelationshipsPretraining2025a} (Sec.~\ref{sec:related}).

\begin{figure}[t]
  \centering
  \includegraphics[width=1.0\linewidth]{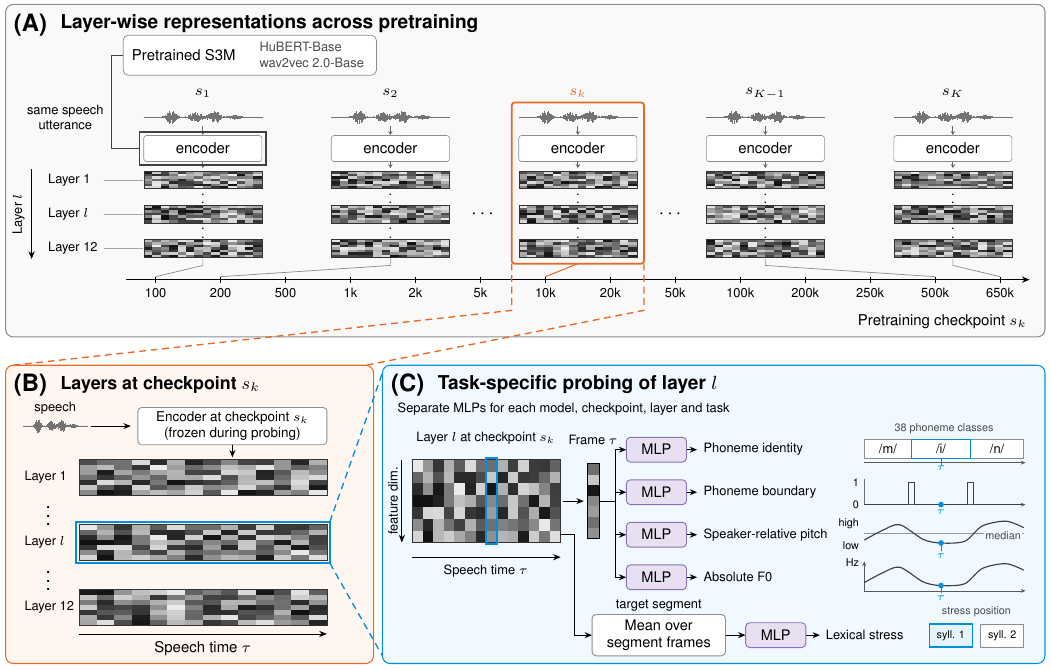}
  \caption{Overview of our probing setup. We extract features at each learning checkpoint (A) and  Transformer layer in S3Ms (B) and use them for downstream tasks. The encoders that produce the input features are frozen during probe training.}
  \label{fig:probing}
\end{figure}

These observations characterize \emph{where} in the network different kinds of
information reside, but say little about \emph{when} such structure arises.
Most layer-wise studies probe a single, fully trained checkpoint, and therefore
cannot reveal how the layer-wise hierarchy emerges or
reorganizes over the course of pretraining. A complementary, smaller
body of work has begun to examine training dynamics: for example, tracking the emergence of
phonetic, lexical, and syntactic structure jointly across layers and
checkpoints of wav2vec~2.0 and HuBERT, and relating each attribute's trajectory
to its degree of abstraction from the acoustic
signal\cite{klootsTrackingEmergenceLinguistic2026b}. Such
reverse-engineering approaches primarily serve as diagnostic tools for
characterizing representation formation; while they may inspire hypotheses
about human language acquisition, any such connection should be interpreted
cautiously and is not direct evidence of cognitive
equivalence\cite{dupouxCognitiveScienceEra2018c}. That work covers phonetic,
lexical, and syntactic structure; what it leaves open is the suprasegmental side
of the picture.

\IEEEpubidadjcol

Prosodic cues are demonstrably encoded by
S3Ms\cite{
fuenteLayerwiseAnalysisMandarin2024, shenEncodingLexicalTone2024,
sunProsodicABXLanguageAgnostic2026}, yet how that localization develops
\emph{relative to} segmental information over training remains poorly
understood. The contrast is also salient in first language acquisition, which
has motivated the use of S3Ms as candidate models of early
learning\cite{dupouxCognitiveScienceEra2018c,
lavechinModelingEarlyPhonetic2024, cruzblandonSimulatingPrenatalLanguage2025b}:
prosody is arguably the earliest information available to the
learner\cite{cruzblandonSimulatingPrenatalLanguage2025b,
gervainProsodyCuesWord2013, mennDoesElectrophysiologicalMaturation2023,
lauInfantPitchPerception2017, lauInfantPitchTimbre2021}, whereas the language-specific
reorganization of phonetic categories unfolds only gradually over the first
year\cite{werkerCrosslanguageSpeechPerception1984a}. 

This asymmetry makes the segmental/suprasegmental split the natural prediction
for a model's acquisition order, and it is that prediction we test, using the infant literature only as a source of hypotheses and claiming no cognitive
equivalence~\cite{dupouxCognitiveScienceEra2018c}. We ask:
\textbf{RQ1)}~In what order do segmental and suprasegmental properties become
decodable over pretraining, and at what depth? \textbf{RQ2)}~Is that order
organized by the linguistic category, or by the degree of abstraction from the
acoustic surface, as proposed by Kloots et
al.~\cite{klootsTrackingEmergenceLinguistic2026b}? \textbf{RQ3)}~Does HuBERT's
pseudo-label target switch leave a signature in layer-wise organization when the
two stages are compared at matched update counts?

We study wav2vec~2.0\cite{baevskiWav2vec20Framework2020}, which optimizes a
single objective throughout, and
HuBERT\cite{hsuHuBERTSelfSupervisedSpeech2021}, whose two stages share an
objective and differ only in their pseudo-label targets. RQ3 is therefore
answered \emph{within} HuBERT, by comparing its two stages at equal update
counts (Sec.~\ref{sec:proposed_approach}). Because wav2vec~2.0 also differs in
objective, it serves as a reference model \emph{without} a target switch.

Our main contributions are:
\begin{itemize}
  \item \textbf{(RQ1)} A five-task design varying abstraction \emph{within} each
        category, so that category and distance from the acoustic surface are
        dissociated. The time-resolved layer-wise lens is that of Kloots et
        al.~\cite{klootsTrackingEmergenceLinguistic2026b}; we add the
        suprasegmental axis on the same checkpoint$\times$layer grid.
  \item \textbf{(RQ2)} A negative result: the order predicted by the
        segmental/suprasegmental split is rejected, the dividing line running
        \emph{within} both categories. This corroborates their abstraction account on a new axis.
  \item \textbf{(RQ3)} A contrast between HuBERT's two stages at equal update
        counts, attributing
        the depth re-localization of phoneme identity to the target change
        rather than to the objective or to longer training, converging with Huo
        and Dunbar~\cite{huoIterativeRefinementNot2025a}.
\end{itemize}

\section{Layer-wise, Time-resolved Analysis}\label{sec:proposed_approach}
We probe a checkpoint$\times$layer grid: at each checkpoint we train a probe on the representations of every Transformer layer, following the setup in Fig.~\ref{fig:probing}; Table~\ref{tab:datasets} lists the five tasks and their data.
HuBERT-Base\footnote{\url{https://github.com/facebookresearch/fairseq/tree/main/examples/hubert}}\cite{hsuHuBERTSelfSupervisedSpeech2021} pairs a CNN feature encoder producing a 20\,ms frame sequence with a 12-layer Transformer that contextualizes a masked version of it.
Following the standard fairseq recipe, HuBERT pretraining proceeds in two stages
with different pseudo-label targets. Stage~1 trains for 250k updates on cluster
IDs from MFCC features; stage~2 then trains a \emph{separately initialized} model for 400k updates on cluster IDs from the 6th Transformer layer of the stage~1 model,
and does \emph{not} continue from the stage-1 checkpoint. All figures plot
\emph{cumulative} updates, so 0--250k is stage~1 and 250k--650k is stage~2's own
0--400k updates. The restart makes stage~1 at 250k updates and stage~2 at 250k
of its own (500k cumulative) step-matched: they share an optimization budget and
an objective and differ only in their targets. This within-model contrast holds
the objective constant.

For comparison, we apply the same probing protocol to
wav2vec~2.0-Base\cite{baevskiWav2vec20Framework2020}, which shares HuBERT's
backbone (a CNN feature encoder and a 12-layer Transformer at a 20\,ms frame
rate) but is trained with a single contrastive objective on masked latent features throughout pretraining, without HuBERT's two-stage pseudo-label switch. Because it also differs in objective, the cross-model comparison cannot
separate objective from curriculum; wav2vec~2.0 therefore serves as a
\emph{reference model without a target switch}. Having no stage split, its
curves show actual updates throughout, so beyond 250k the two models sit at
different update counts at the same horizontal position.

We save checkpoints on a log-spaced schedule (from 100 steps to 650k),
sampling densely in early pretraining and progressively coarser toward
convergence, since representations reorganize fastest at the start and
change little once the probing metrics saturate. Step counts are
\emph{cumulative} across HuBERT's two stages, so the schedule places a
checkpoint at the end of stage~1 (250k cumulative) and at 500k and 650k cumulative within stage~2. We use a multi-layer perceptron (MLP) probe trained on frozen representations, with a classification head for all tasks except the
frame-level F0 task, which uses a regression head.\vspace{1em}

\begin{table}[t]
  \centering
  \caption{Datasets \& duration time}
  \label{tab:datasets}
  \begin{tabular}{lll}
    \hline
    Task & Dataset & Duration \\
    \hline
    Pitch & PTDB-TUG & 9.6 hours \\
    F0 & PTDB-TUG & 9.6 hours \\
    Phoneme boundary & TIMIT & 4.3 hours \\
    Phoneme & TIMIT & 4.3 hours \\
    Lexical Stress & Prosody-ABX & 4.6 minutes \\
    \hline
  \end{tabular}
\end{table}

\noindent\textbf{Pitch task}
To probe pitch height relative to a speaker's own range, we use PTDB-TUG\cite{pirkerPitchTrackingCorpus2011} (9.6\,h), in which native English speakers read TIMIT sentences\cite{garofolojohns.TIMITAcousticPhoneticContinuous1993a}; its laryngograph-derived F0 reference (100 Hz) is nearest-neighbour resampled to the model frame rate (50 Hz). Unvoiced frames are discarded, and each remaining frame is labelled high or low according to whether its log-F0 exceeds the median log-F0 over that speaker's voiced frames. The label is a pitch height at one point in time, not a change between time points, and the per-speaker threshold cancels the speaker's pitch register. A held-out-speaker split (6 of 20) prevents the probe from memorising those thresholds.\vspace{1em}

\noindent\textbf{F0 regression task}
To probe absolute pitch, we run frame-level F0 regression on PTDB-TUG following \cite{fuenteLayerwiseAnalysisMandarin2024}, using voiced frames only and feeding the per-layer frame embedding directly into the probe, without temporal pooling, to predict the laryngograph-derived reference F0 in Hz under the same speaker split.\vspace{1em}


\noindent\textbf{Phoneme boundary task}
 We frame phone-segment boundary detection as a frame-level binary task on TIMIT (4.3\,h). Reference boundaries come from the TIMIT phone alignments, mapped to the model's 50 Hz frame grid, with utterance-initial and -final edges excluded. Strict boundary frames are the positives (tolerance 0). As for the other tasks, the utterance is encoded in a single forward pass and the per-layer frame features are used directly. Because exact frame-level localization is overly strict on a 20 ms grid, we report a tolerant boundary F1: predicted and reference boundaries are matched one-to-one within $\pm1$ frame ($\pm20$ ms) by greedy nearest-matching within each utterance.\vspace{1em}

 \noindent\textbf{Phoneme classification task}
 TIMIT\cite{garofolojohns.TIMITAcousticPhoneticContinuous1993a} provides manually time-aligned phoneme segmentations for 630 speakers across eight US dialects. We use the SI and SX sentences (excluding the two dialect ``SA'' sentences) under the standard train/test split (3696/1344 utterances). Following the reduced phoneme set of Lee and Hon \cite{lee1989speaker}, we map the 61 TIMIT labels to the standard 39-phone set and use 38 phoneme classes, excluding silence.\vspace{1em}

\noindent\textbf{Lexical stress task}
Using the English-Stress subset of the Prosody-ABX corpus\cite{sunProsodicABXLanguageAgnostic2026}, each clip is a target word whose stressed-syllable index a native annotator marks; we treat this as binary classification of stress position, mean-pooling the per-layer frame features over the segment. Because the corpus is small (600 segments, 10 speakers, 4.6\,min), we use speaker-grouped 5-fold cross-validation (each speaker in exactly one test fold) and report the mean test macro-F1.\vspace{1em}

\subsection{Experimental setup}
Both models are pretrained from scratch with Fairseq\cite{ottFairseqFastExtensible2019} on LibriSpeech 960h\cite{7178964} on two A100 80GB GPUs, with an effective batch of 4.4M audio samples for HuBERT and 16.8M for wav2vec~2.0, the larger being needed for stable training. 

The probe is an MLP with a single hidden layer of 256 units, ReLU and dropout
$0.1$, trained with Adam (constant learning rate, weight decay $10^{-4}$) and
early stopping on a held-out 10\% of the training data. The learning rate is
$3{\times}10^{-4}$ ($10^{-3}$ for boundary) and the batch size 2048 (256 for
stress); patience is 3 epochs for the TIMIT tasks and 5 for PTDB-TUG and stress,
capped at 50 epochs (60 and 100 respectively). Phoneme and boundary probes use
balanced class weights, and no speaker embedding is concatenated.
Probes are evaluated with F1, except F0 regression ($R^2$). Our analyses target the learning dynamics of each attribute, when and where it becomes decodable; absolute performance is not comparable across models given their different objectives and configurations.

As a capacity control, we also train a linear probe on the phoneme task under
the same protocol with the hidden layer removed. Its best layer is L5 at 250k
and L9 at 500k and 650k, exactly as for the MLP, and the re-localization across
depth is larger without the hidden layer: from 250k to 650k, mean macro-F1 over
L10--L12 rises $0.56\!\to\!0.72$ (MLP $0.61\!\to\!0.74$) while L1--L3 stay flat
in both. The same shift therefore appears without any nonlinear probe capacity.


\section{Results}

\begin{figure}[!t]
  \centering
  \includegraphics[
  width=\columnwidth,
  height=0.35\textheight,
  keepaspectratio
  ]{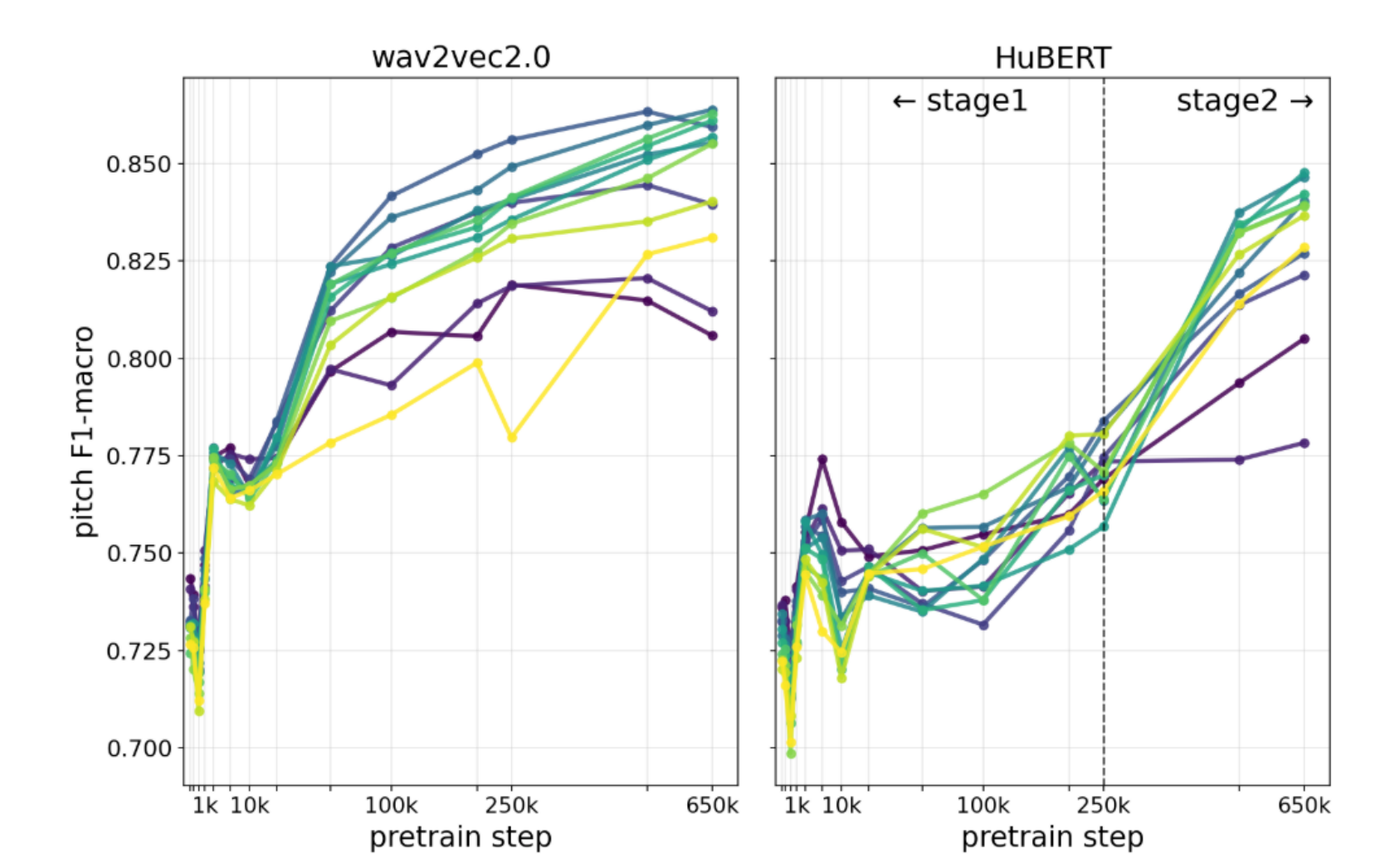}
  \caption{Probe accuracy for speaker-relative pitch. $x$: log-spaced
  \emph{cumulative} updates; for HuBERT, 0--250k is stage~1 and 250k--650k the
  separately initialized stage~2; wav2vec~2.0 has no stage split.}
  \label{fig:pitch}
\end{figure}
\noindent\textbf{Pitch task}
The pitch task is broadly and shallowly encoded and improves only gradually over pretraining. Best-layer F1 rises from $\approx0.74$ at 100 steps to $0.85$--$0.86$ at 650k, and the two models differ mainly in pace: wav2vec~2.0 is essentially saturated by 250k ($0.856$), whereas HuBERT's stage-2 run ends higher than stage~1 ($0.78$ at 250k $\to$ $0.85$ at 650k). Its layer profile is the flattest of all tasks: at convergence all twelve layers lie within $\approx0.80$--$0.86$, with only a mild middle-layer preference (L5 for wav2vec~2.0, L7 for HuBERT). Unlike absolute F0, which peaks sharply in the lowest layers, speaker-relative pitch peaks mid-network and is spread evenly across depth.\vspace{1em}

\begin{figure}[h]
  \centering
  \includegraphics[
  width=\columnwidth,
  height=0.35\textheight,
  keepaspectratio
  ]{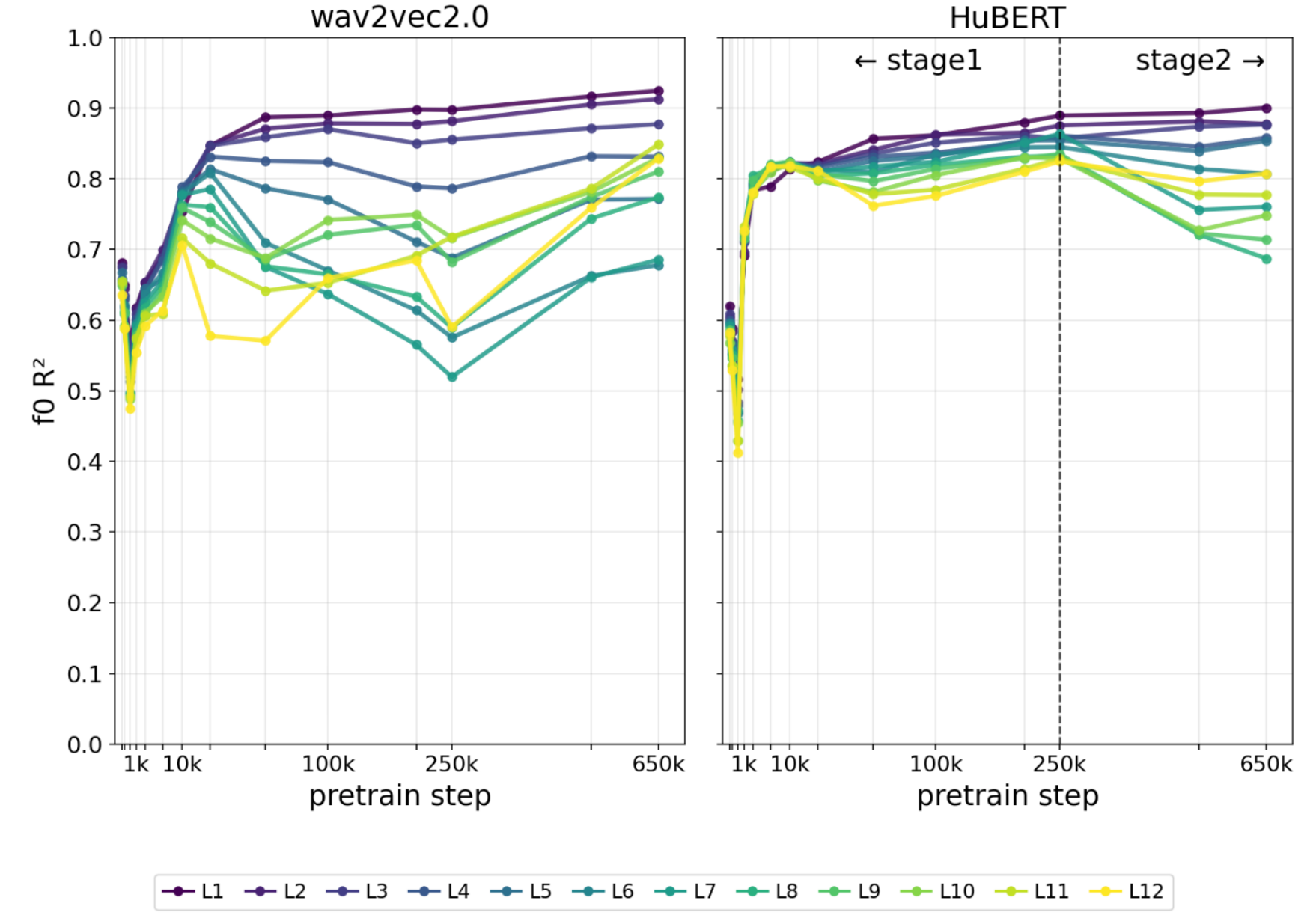}
  \caption{F0 regression performance ($R^2$). Axis as in Fig.~\ref{fig:pitch}.}
  \label{fig:f0_regression}
\end{figure}
\noindent\textbf{F0 regression task}
Absolute F0 is recoverable almost from the start of pretraining and remains a low-layer property throughout. The lowest layers (L1--L2) reach $R^2\approx0.82$ by 10k and climb to $0.89$ (HuBERT) and $0.92$ (wav2vec~2.0) by 650k, and at every checkpoint the best layer is L1. Across depth the profile forms a U-shape that deepens over training: the low layers stay near ceiling while the middle layers move away from the raw pitch value. For HuBERT, all twelve layers are still tightly clustered at the end of stage~1 ($\approx0.82$--$0.89$ at 250k), whereas under the stage-2 targets the middle and upper layers fall away (mean L5--L8 $0.85\to0.78$, minimum $\approx0.69$ at L8); these endpoints differ in update count (250k vs.\ 400k) as well as in target, and the no-switch wav2vec~2.0 develops the same divergence earlier and more strongly (mid layers $0.68$--$0.77$ against $\approx0.92$ in the low layers), so the divergence is not specific to the target change. In both models the outcome is a clear division of labour along depth: raw F0 stays best recovered in the low layers, while the middle and upper layers increasingly re-encode pitch into a more abstract form that is less aligned with the absolute F0 value.\vspace{1em}

\begin{figure}[!t]
  \centering
  \includegraphics[
  width=\columnwidth,
  height=0.35\textheight,
  keepaspectratio
    ]{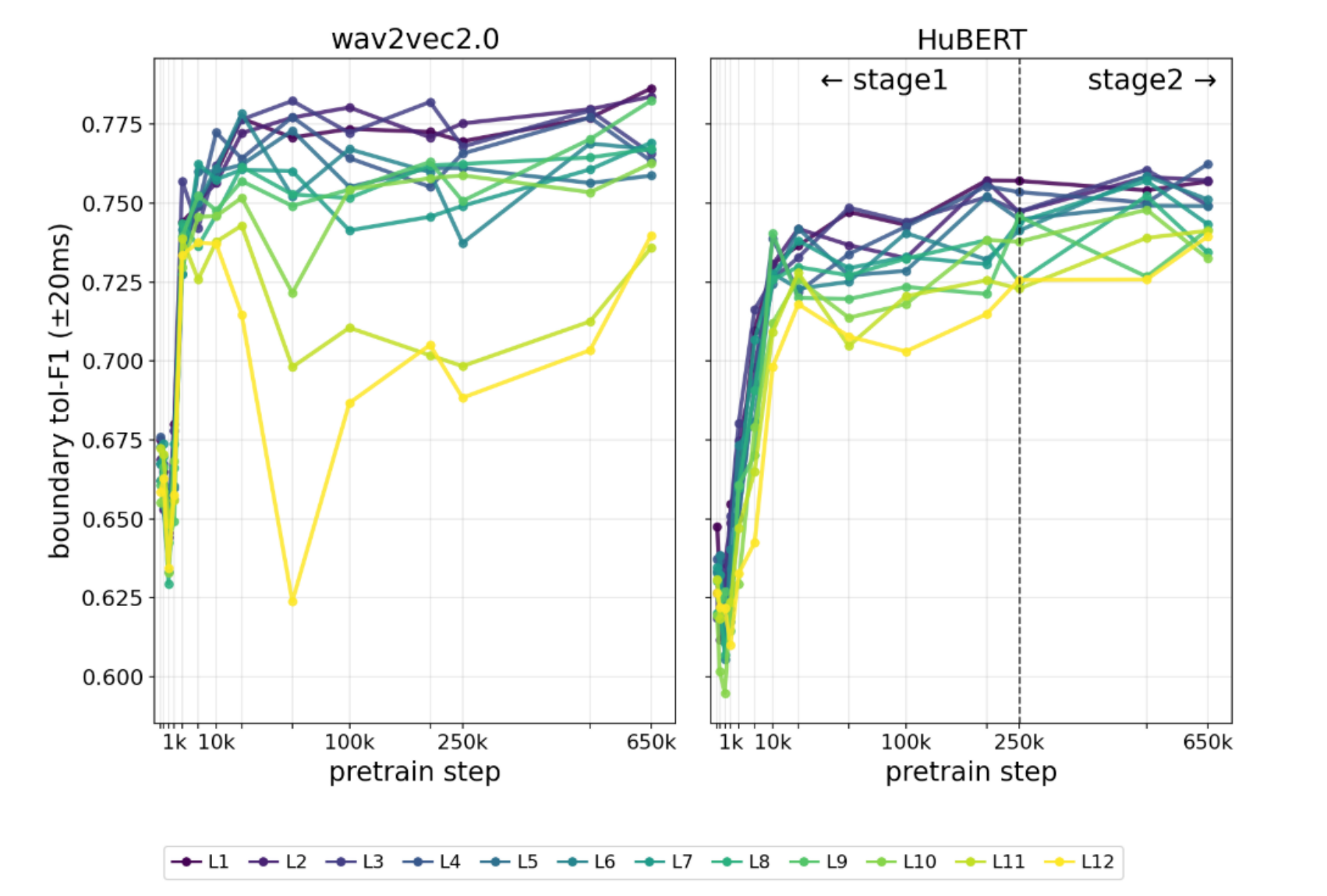}
  \caption{Probe accuracy for phoneme boundary detection. Axis as in Fig.~\ref{fig:pitch}.}
  \label{fig:phoneme_boundary}
\end{figure}

\noindent\textbf{Phoneme boundary task}
Boundary detection is markedly different: it is largely established within the first few thousand steps and nearly flat thereafter, across both training and depth. Best-layer tolerant F1 ($\pm20$ ms) is already $0.65$--$0.68$ at 100 steps and rises only to $0.76$ (HuBERT) and $0.79$ (wav2vec~2.0) at 650k. At convergence the profile is nearly uniform (all twelve layers within $\approx0.73$--$0.79$) and the best layer sits unusually early (L4 for HuBERT, L1 for wav2vec~2.0). The only systematic change is a catch-up of the upper layers (L10--L12, $\approx0.63$--$0.66$ to $\approx0.74$--$0.75$). Boundary cues thus rest on local acoustic change, available shallowly and early and not sharpened much by further pretraining.\vspace{1em}

\begin{figure}[!t]
  \centering
  \includegraphics[
    width=\columnwidth,
    height=0.35\textheight,
    keepaspectratio
  ]{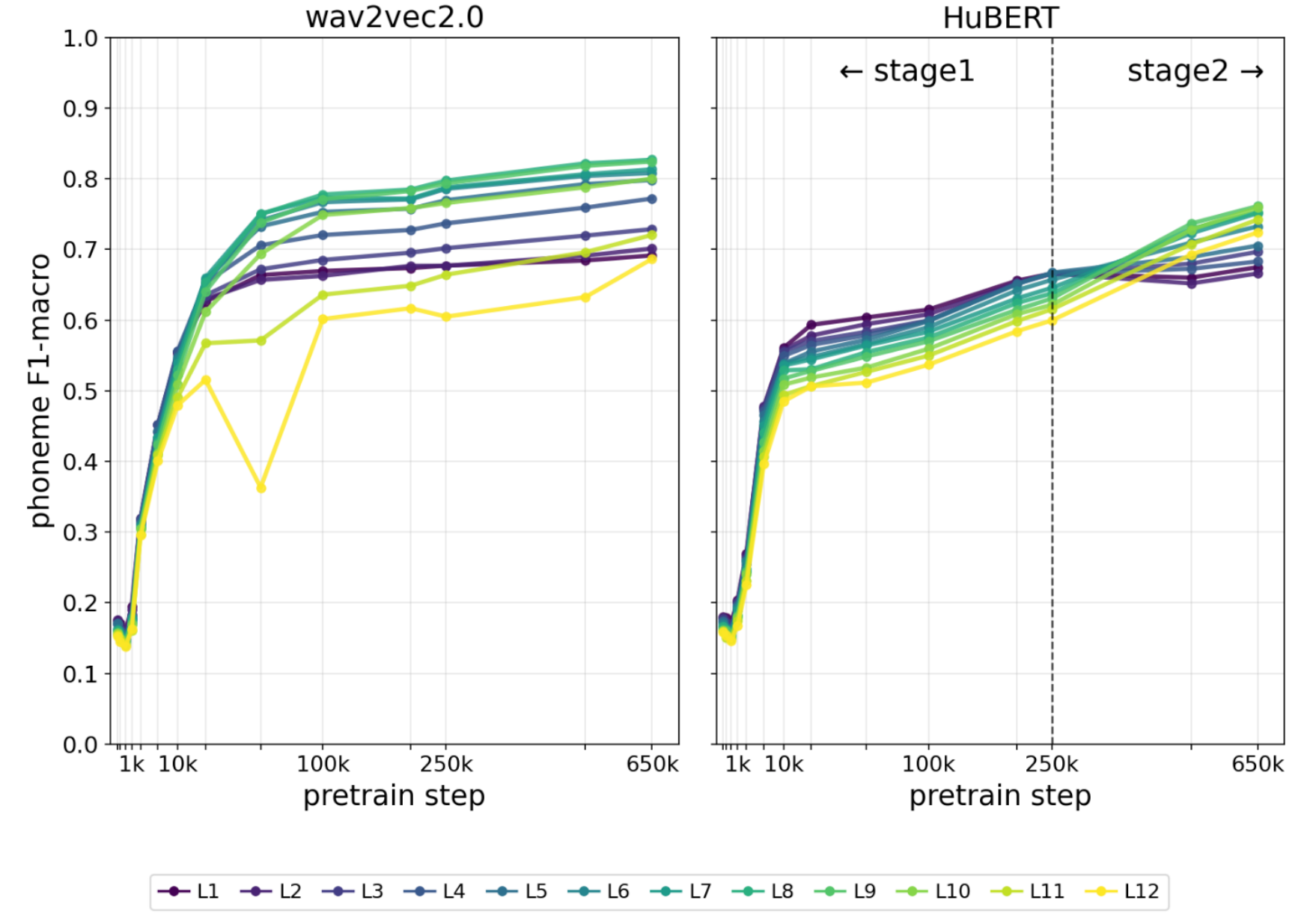}
  \caption{Probe accuracy for phoneme task. Axis as in Fig.~\ref{fig:pitch}.}
  \label{fig:phoneme_probe}
\end{figure}

\begin{figure}[!t]
  \centering
  \includegraphics[
    width=\columnwidth,
    height=0.35\textheight,
    keepaspectratio
  ]{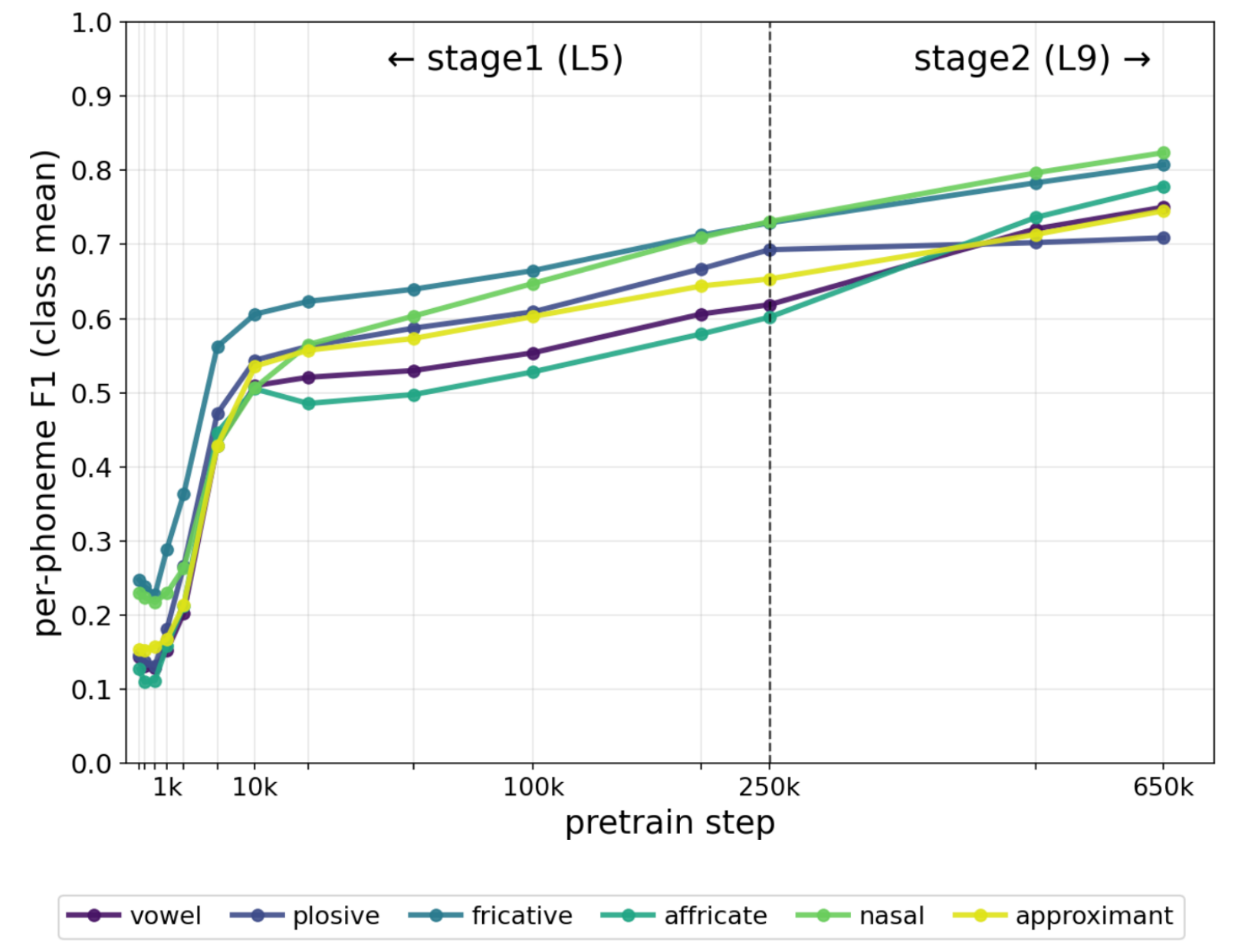}
  \caption{Per-phoneme-class F1 across pretraining. Axis as in Fig.~\ref{fig:pitch}.}
  \label{fig:phoneme_comparison}
\end{figure}

\noindent\textbf{Phoneme classification task}
Phoneme discrimination is almost entirely built by pretraining. Best-layer macro-F1 starts at $\approx0.18$ at 100 steps for both models and rises steeply: wav2vec~2.0 reaches $0.75$ by 50k and $0.83$ at 650k, whereas HuBERT rises more gradually ($0.48$ at 5k, $0.60$ at 50k, $0.76$ at 650k). The layer profile also changes over training. At the earliest steps the best layer is L1, but the peak moves upward: for HuBERT it reaches L5 by the end of stage~1 (250k) and to L9 by 500k, so that by 650k a clear mid-to-upper peak has formed (L7--L10 $\approx0.75$) while the lowest layers plateau (L1--L3 $\approx0.68$) and the final layers fall back (L11--L12 $\approx0.72$). Between 250k and 650k the middle and upper layers surge (mean L5--L8 $0.65\!\to\!0.74$; L10--L12 $0.61\!\to\!0.74$) while the low layers stay flat ($0.67\!\to\!0.68$). wav2vec~2.0 differs: its best layer settles at L7--L8 by ${\sim}20$k and stays, and at convergence its top two layers are weakest (L11--L12 $0.72/0.69$ vs.\ L8 $0.83$).\vspace{1em}

\noindent\textbf{Per-phoneme development}
Figure~\ref{fig:phoneme_comparison} tracks HuBERT's per-phoneme F1, averaged within phonetic class, over log-spaced pretraining steps, using each stage's best layer (L5 for checkpoints up to 250k and L9 afterwards); read this way the two runs meet at 250k with no visible step. Acquisition is strongly front-loaded: every class jumps from near chance at $\sim$100 steps ($0.13$--$0.25$) to $0.5$--$0.6$ by 10k. Improvement then slows: classes gain $+0.47$ to $+0.55$ F1 over the whole of stage~1 (to 250k) but end only $+0.02$ to $+0.18$ higher at the stage-2 endpoint (650k) than at the end of stage~1, confirming that most phoneme-discriminative structure is laid down early in the lower-middle layers, with the stage-2 model differing mainly in smaller upper-layer refinements.

The ranking also changes. Fricatives lead throughout stage~1 ($0.73$ at 250k, tied with nasals), whereas vowels and affricates begin lowest; by 650k nasals ($0.82$) and fricatives ($0.81$) remain on top, but affricates gain most from stage~1 to stage~2 ($+0.18$, to $0.78$), ahead of vowels and approximants (both $0.75$), while plosives, among the strongest early ($0.69$ at 250k), are almost unchanged ($+0.02$) and end lowest ($0.71$). Context-dependent categories thus benefit from the upper-layer contextualization the stage-2 model develops, whereas fricatives and nasals are captured from local acoustics early on.\vspace{1em}

\begin{figure}[!t]
  \centering
  \includegraphics[
  width=\columnwidth,
  height=0.35\textheight,
  keepaspectratio
  ]{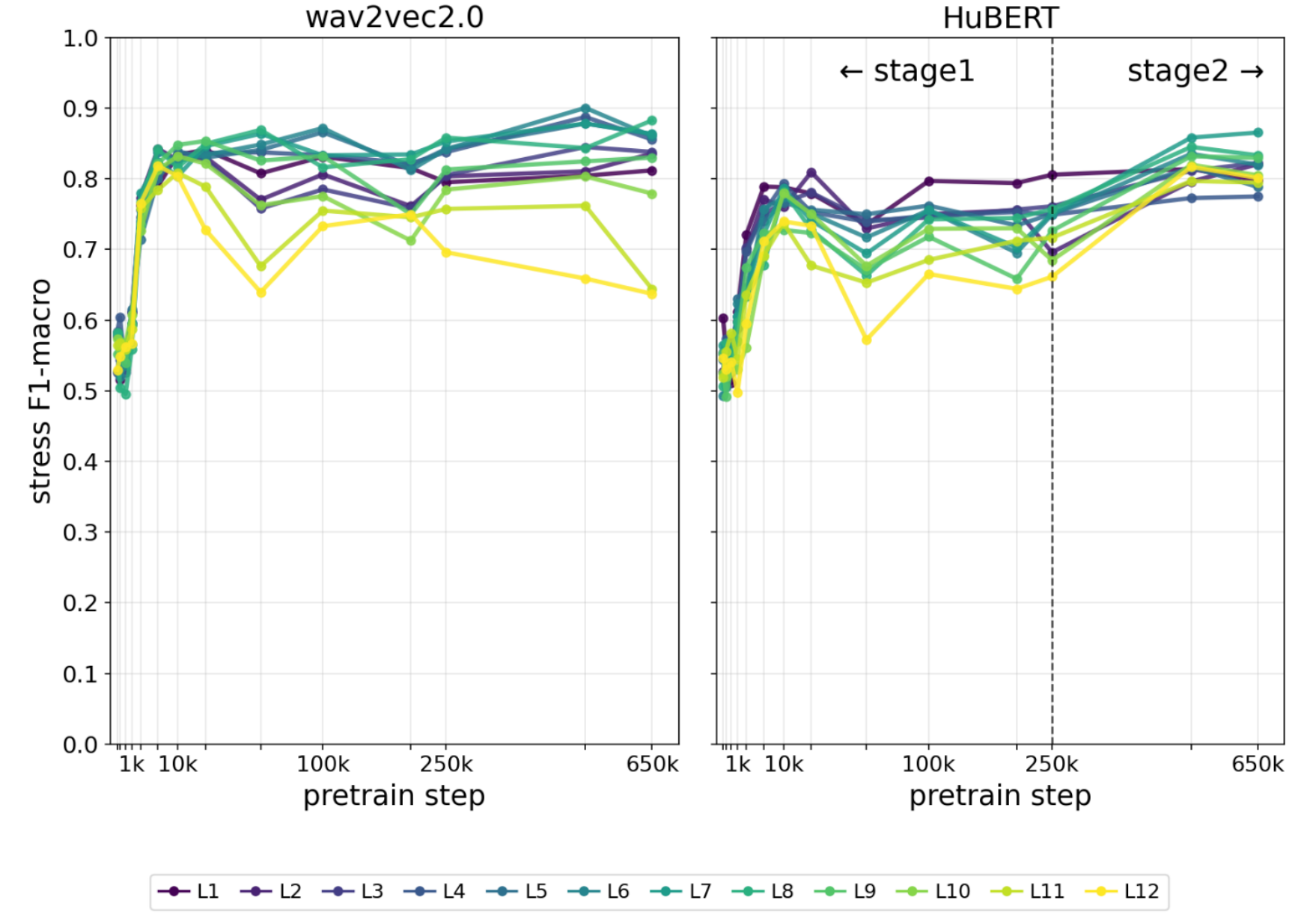}
  \caption{Probe accuracy for stress ABX. Axis as in Fig.~\ref{fig:pitch}.}
  \label{fig:stress_abx}
\end{figure}

\noindent\textbf{Lexical stress task}
Lexical stress is weakly present at initialization and emerges over early pretraining, but the models differ in pace. Best-layer F1 starts modestly above chance at 100 steps ($0.60$ for HuBERT, $0.58$ for wav2vec~2.0); wav2vec~2.0 essentially saturates by 50k ($0.87$, reaching $0.88$ at 650k), whereas HuBERT continues to improve well beyond 50k ($0.75$ at 50k $\to$ $0.865$ at 650k, i.e.\ still $+0.12$ after 50k). Layer differentiation is small: a mild mid-layer preference at convergence (L7 for HuBERT, L8 for wav2vec~2.0) and no re-localization comparable to the phoneme task. The main depth effect is a drop in the final layers, pronounced for wav2vec~2.0, whose top two fall to $0.64$ (vs.\ $0.88$ at L8): stress, like phoneme identity, is pushed out of the topmost layers as they specialize toward the objective.

\section{Discussion}

\noindent\textbf{Phoneme information is re-localized across depth, not uniformly sharpened.}
For the phoneme task (Fig.~\ref{fig:phoneme_probe}), what changes most clearly between the two stages is \emph{where} phoneme identity is decodable: in stage~1 the lower layers are most decodable, under the stage-2 targets the upper-middle ones. The evidence is the re-localization across depth rather than the best-layer shift alone: between 250k and 650k, mean macro-F1 rises $0.65\!\to\!0.74$ over L5--L8 and $0.61\!\to\!0.74$ over L10--L12 while L1--L3 stay flat ($0.67\!\to\!0.68$), whereas a uniform sharpening would have lifted all depths. Those two checkpoints differ in optimization budget (250k vs.\ 400k updates), but the step-matched pair points the same way: after 250k of stage~2's own updates the peak already sits at L9, and the linear probe reproduces the same move. F1 is still rising at 650k, so decodability has not saturated within our budget, and because the tasks differ in chance level and number of classes the large phoneme gain partly reflects headroom; the re-localization, which the near-ceiling tasks do not exhibit, does not. The gain is thus confined to particular depths: pretraining changes where phoneme identity can be read out, not only how well.\vspace{1em}

\noindent\textbf{The segmental/suprasegmental prediction fails: acquisition order tracks abstraction from the acoustic surface.}
The tasks differ systematically in when and where each property is encoded: absolute F0 is present from the outset in the low layers, stress saturates early in wav2vec~2.0, speaker-relative pitch and HuBERT's stress are refined gradually, and phoneme identity improves throughout pretraining. The organizing variable is therefore not the linguistic category but the degree of abstraction from the acoustic signal, as proposed by Kloots et al.~\cite{klootsTrackingEmergenceLinguistic2026b}. Five tasks varying abstraction \emph{within} each category put the competing, category-based prediction at risk. Cues readable off the acoustic surface (F0, stress, and phone \emph{boundaries}, a segmental property carried by local spectral change) are available early and shallow, whereas phoneme \emph{identity}, which requires context, is constructed late and deep. Two \emph{segmental} properties thus follow opposite trajectories, so the category label alone does not predict the dynamics. Across all tasks the low layers remain acoustically fixed while the middle-to-upper layers differentiate; the middle layers move away from raw F0 yet toward phoneme identity and speaker-relative pitch, marking them as the acoustic-to-abstract conversion point of the network.\vspace{1em}


\noindent\textbf{Target change, not training objective, drives the re-localization.}
HuBERT and wav2vec~2.0 differ not only in whether the target changes but also in the objective itself (masked-prediction classification vs.\ a contrastive loss with negatives), so the cross-model contrast cannot separate curriculum from objective; wav2vec~2.0 is only a reference model \emph{without} a target switch. The clean evidence is internal to HuBERT, where the objective is identical across stages and only the targets change: across the step-matched pair the phoneme peak sits at L5 in stage~1 but at L9 in stage~2, so the re-localization follows the target change, not the objective or the training duration. Huo and Dunbar\cite{huoIterativeRefinementNot2025a} converge on this, finding that iterative pseudo-label refinement rather than the objective accounts for the differences between the two models; we add its trajectory across depth and training time. Upper-layer phoneme decodability rises across the two stages while speaker-relative pitch decodability does not (Figs.~\ref{fig:pitch},~\ref{fig:phoneme_probe}), a suggestion of prosodic-vs-phonemic differentiation across depth; at the final checkpoint our layer-wise trends are broadly consistent with prior analyses of fully trained models~\cite{pasadLayerwiseAnalysisSelfsupervised2022}.\vspace{1em}

\noindent\textbf{Limitations.}
Our stage-2 checkpoints are sparse (500k and 650k cumulative, against the 250k stage-1 endpoint), so we compare the stages at matched update counts but cannot localize \emph{when} within stage~2 the re-localization occurs; denser ones are left to future work. The stress set is small (600 segments, 10 speakers, 4.6~min): speaker-grouped cross-validation guards against speaker memorization but not against the narrow speaker, lexical and language coverage, so that trajectory is the least generalizable of our results; the pipeline of~\cite{bentumWordStressSelfsupervised2025} scales better. We also lack an architecture-matched non-speech control: at 100 steps a near-untrained encoder already reaches most of the final boundary and speaker-relative pitch scores (phoneme identity is still $\approx0.18$), so how much of this reflects architecture and input statistics remains open. Finally, all data are English read speech from a single run per model, and best
layers are single-seed argmaxes (at 250k the margin is under $0.002$), so our claims rest on the layer-group contrasts.

\section{Related Work}\label{sec:related}

\subsection{Layer-wise organization of speech information in S3Ms}
Probing classifiers~\cite{belinkovProbingClassifiersPromises2022} measure how
decodable an attribute is from frozen representations, and applied across depth
they have established a robust hierarchy in S3Ms: acoustic and speaker cues
dominate the low layers, phonetic and word-level information peaks in the
middle, and the top layers re-specialize toward the pretraining
objective~\cite{pasadLayerwiseAnalysisSelfsupervised2022,pasadComparativeLayerwiseAnalysis2023b,chungSimilarityAnalysisSelfSupervised2020};
geometric analyses further place speaker and phonetic information in nearly
orthogonal
subspaces~\cite{liuSelfsupervisedPredictiveCoding2023,mohamedOrthogonalityIsotropySpeaker2024}.
Our profiles reproduce this hierarchy at 650k, but these studies characterize a
single fully trained checkpoint and are silent on \emph{when} it forms. Our checkpoint-resolved view shows it is not one schedule: the low-layer acoustic profile is in place within a few thousand updates, the phoneme peak only late.

\subsection{Learning dynamics across pretraining}
A smaller body of work probes representational change across training. In NLP, \cite{inabaHowBilingualLM2025} track how bilingual alignment
strengthens across steps and layers, while Kloots et al.~\cite{klootsTrackingEmergenceLinguistic2026b} follow the
emergence of phonetic, lexical, and syntactic structure jointly across layers and checkpoints of wav2vec~2.0 and HuBERT, relating each
attribute's trajectory to its degree of abstraction from the acoustic signal.
Building on their time-resolved, layer-wise lens and their abstraction account,
we add
(i)~the \emph{suprasegmental} axis on the same checkpoint$\times$layer grid as
the segmental tasks, which makes the category-based prediction testable at all,
and (ii)~a comparison of HuBERT's two stages that holds the objective constant in a way the cross-model comparison cannot.
Huo and Dunbar~\cite{huoIterativeRefinementNot2025a} reach a compatible
conclusion by attributing the differences between the two models to iterative pseudo-label refinement.

\subsection{Prosody in S3Ms and links to language acquisition}
Probing and discrimination studies recover stress, pitch accent, and lexical
tone from intermediate layers~\cite{fuenteLayerwiseAnalysisMandarin2024,shenEncodingLexicalTone2024}, and
training-free measures such as prosodic ABX~\cite{sunProsodicABXLanguageAgnostic2026} show that
prosodic contrast is encoded language-agnostically, with a layer-wise profile
distinct from that of phonemic contrast. These profiles are read off trained models; our checkpoint-resolved view shows they are not reached by a common route: absolute F0 is recoverable from the lowest layers almost from the start, whereas speaker-relative pitch peaks mid-network and is refined gradually, so ``prosody'' is not one trajectory. Closest to our stress probe, Bentum et
al.~\cite{bentumWordStressSelfsupervised2025} find word-stress representations
in wav2vec~2.0 XLS-R to be language-specific across five languages, evaluating
trained models on ${\sim}10^{5}$ automatically labelled tokens per language,
whereas we track stress across pretraining. This segmental/suprasegmental split is
also central to first language acquisition, which has motivated treating S3Ms
as candidate learning
systems~\cite{dupouxCognitiveScienceEra2018c,lavechinModelingEarlyPhonetic2024,lavechinReverseEngineeringLanguage}: infant pitch
perception is functional within the first months of
life~\cite{lauInfantPitchPerception2017,lauInfantPitchTimbre2021}, whereas native phonetic categories are
reorganized more gradually over the first
year~\cite{werkerPhoneticPerceptualReorganization2024a,werkerCrosslanguageSpeechPerception1984a,feldmanInfantsReallyLearn2021}. The acquisition literature supplies the hypothesis our five tasks evaluate and reject; we use it
only as a source of hypotheses about acquisition \emph{order} and make no claim
of cognitive equivalence~\cite{dupouxCognitiveScienceEra2018c,feldmanInfantsReallyLearn2021}.

\section{Conclusion}
We tracked how segmental (phoneme boundary and identity) and suprasegmental
(absolute F0, speaker-relative pitch, lexical stress) information emerges in HuBERT and
wav2vec~2.0, across all layers and the full pretraining schedule. Acquisition
order follows the degree of abstraction from the acoustic surface rather than
the segmental/suprasegmental distinction: cues carried by local spectral structure (F0, stress, phone boundaries) are decodable early in the low layers, whereas phoneme identity is built throughout training and re-localizes
to the upper-middle layers under HuBERT's stage-2 targets. That order rejects the category-based prediction and extends onto the suprasegmental axis an existing abstraction account so far established for phonetic, lexical and syntactic structure. Within HuBERT, the two stages at equal update counts differ only in targets, attributing that
re-localization to the target switch rather than to the objective or to longer
training.

\section*{Acknowledgment}
GPT-5/5.4/5.5 and Claude-Fable/Opus/Sonnet were used to assist with parts of
the analysis and figure-generation code, and to help draft and revise the
manuscript. The authors verified all code, results and text, and take full
responsibility for the content.

This work was supported by JSPS KAKENHI Grant Numbers JP26H02537, JP25KJ1833, and JP25H01139, and JST NEXUS Japan Grant Number JPMJNX25C1.

\bibliographystyle{IEEEtran}
\bibliography{refs_tanibata}
\end{document}